%% file: main.tex
\documentclass[sigplan]{acmart}

\usepackage{amsmath}
\usepackage{amsfonts}
\usepackage{amsthm}
\usepackage{caption}
\usepackage{appendix}
\usepackage[T1]{fontenc}
\usepackage{footmisc}
\usepackage{graphicx}
\usepackage[utf8]{inputenc}
\usepackage{nicefrac}
\usepackage{soul}
\usepackage{subcaption}
\usepackage{titlesec}
\usepackage{url}
\usepackage{verbatim}
\usepackage{xcolor}

\definecolor{darker}{rgb}{0,0.15,0.8}
\definecolor{darkred}{rgb}{0.7,0.0,0.0}

\renewcommand\footnotetextcopyrightpermission[1]{}

\begin{document}

\title[Deep Learning Computational Challenges]{Beyond Human-Level Accuracy: Computational Challenges in Deep Learning}


\author{Joel Hestness}
\author{Newsha Ardalani}
\author{Gregory Diamos}
\affiliation{
  \institution{Baidu Research}
  \streetaddress{1195 Bordeaux Dr.}
  \city{Sunnyvale}
  \state{California}
  \postcode{94089}
  \country{USA}
}
\email{jthestness@gmail.com}
\email{newsha@baidu.com}
\email{gregory.diamos@gmail.com}

\copyrightyear{2019}
\acmYear{2019}
\setcopyright{none}
\acmConference[arXiv]{arXiv draft}{August 19, 2019}{Sunnyvale CA}
\acmPrice{15.00}
\acmDOI{10.1145/3293883.3295710}
\acmISBN{978-1-4503-6225-2/19/02}

\input{abstract}

\maketitle


\input{introduction}

\input{applications}

\input{app_scaling}

\input{compute_scale}

\input{projections}

\input{case_study}

\input{related}
\input{conclusion}

\bibliography{bibliography}

\newpage
\appendix
\input{appendices}

\end{document}

%% file: abstract.tex
\begin{abstract}

Deep learning (DL) research yields accuracy and product improvements from both model architecture changes and scale: larger data sets and models, and more computation. For hardware design, it is difficult to predict DL model changes. However, recent prior work shows that as dataset sizes grow, DL model accuracy and model size grow predictably. This paper leverages the prior work to project the dataset and model size growth required to advance DL accuracy beyond human-level, to frontier targets defined by machine learning experts. Datasets will need to grow $33$--$971\times$, while models will need to grow $6.6$--$456\times$ to achieve target accuracies.

We further characterize and project the computational requirements to train these applications at scale. Our characterization reveals an important segmentation of DL training challenges for recurrent neural networks (RNNs) that contrasts with prior studies of deep convolutional networks. RNNs will have comparatively moderate operational intensities and very large memory footprint requirements. In contrast to emerging accelerator designs, large-scale RNN training characteristics suggest designs with significantly larger memory capacity and on-chip caches.


\end{abstract}

%% file: introduction.tex
\section{Introduction}
\label{sec:introduction}

Deep learning (DL) has emerged as a primary driver of recent artificial intelligence (AI) breakthroughs. As DL-enabled products grow, it becomes more important to satisfy the future hardware requirements of DL model training. We aim to develop systems that meet these future DL requirements.

DL accuracy advances can come from two different drivers. First, DL researchers study model architecture changes to better fit data sets and improve accuracy. Model changes tend to be highly non-trivial---often requiring problem reframing---and can substantially change their computational structure. As a result, it is very difficult to predict the model structures that will be important for future DL applications.

However, DL advances can also come from growing data sets, model size, and computation---an approach that has received more attention in recent research. The DL community commonly accepts that model accuracy improves as training dataset size grows (e.g., \cite{banko:verylargenld:acl:2001,amodei:ds2:icml:2016,sun:dataeffective:iccv:2017}). Further, Hestness et al. characterize accuracy and model size growth, showing they are particular power-law functions of dataset size \cite{hestness:dlpredictscaling:arxiv:2017}.

This paper leverages the prior work to project the data and model size scaling required to advance DL accuracy beyond human-level, to frontier targets defined by machine learning experts. We collect these accuracy targets for five DL domains---word and character language modeling, machine translation, speech recognition, and image classification.

These domains will require substantial increases in dataset and model size to achieve target accuracy. Datasets will need to grow in size $33$--$971\times$ larger than the datasets used to train current state-of-the-art (SOTA) models. Models must also grow in parameter count by $6.6$--$456\times$ larger. Based on these desired targets, simple estimates suggest that training time would take decades to centuries on current systems.

Not shying away from the challenge, this paper characterizes and projects the growth in computational requirements to train these target applications. Although some DL applications are computationally well-understood, our broader analysis reveals surprisingly predictable compute and memory scaling across a range of very different DL architectures, including deep convolutional networks (CNNs), recurrent sequence-to-sequence models, and recurrent encoder-decoder models with attention.

Our characterization reveals an important segmentation of DL training challenges. While prior works have focused heavily on CNNs, their compute requirements differ significantly from recurrent neural networks (RNNs), which are likely to demand far more compute and memory resources. Image processing applications with deep CNNs desire relatively small growth in dataset and model size, and they show more potential to leverage emerging compute accelerators with high compute-to-memory throughput ratios. Even small batch sizes can expose sufficient operational intensity for high compute throughput.

On the other hand, RNNs, especially in language domains, will require upwards of $100\times$ more training time to achieve target accuracy. They have moderate operational intensities, and very large memory footprints that exceed current accelerator memory capacity by $8$--$100\times$. These characteristics make it difficult to efficiently parallelize large-scale training, as we demonstrate in a language modeling case study.

We recommend the hardware community place more focus on supporting RNNs. Systems for RNN training could be substantially different than emerging hardware. For example, a possible approach to better support large-scale RNN training parallelism would be to significantly increase accelerator memory capacity. We could also better leverage growing accelerator compute throughput by building larger on-chip caches to avoid excessive memory data streaming for large matrix multiply operations. These approaches run counter to emerging accelerator designs.

%% file: applications.tex
\section{Deep Learning Applications}
\label{sec:applications}

The deep learning research community has developed a large set of important DL applications. The initial release of MLPerf \cite{mlperf:benchmark:2018} identifies seven domains critical for industry DL training: image classification and object detection, recommendations, reinforcement learning in games, language understanding sentiment analysis, speech recognition, and translation. This paper focus on some MLPerf applications and a similar breadth: image classification, language modeling, speech recognition, and translation. This section describes the general algorithmic structure of DL applications, and describes the particular applications for which we study scaling behaviors.

%
\subsection{Compute Graphs of DL Applications}

Deep learning applications are usually structured algorithmically as compute graphs. These compute graphs include nodes, or "ops", that perform a mathematical computation---e.g., matrix-vector multiplication, convolution, or pointwise operations---on input data. Boxes in the network diagrams below represent ops or groups of ops. Data is passed between ops using "tensors" (like data arrays) that encode the data's structure and dependencies between ops.

To project future hardware needs, we define three properties of the compute graphs that allow us to characterize their compute and memory requirements. In practice, when executing a compute graph on hardware, numerous hardware factors affect performance and are difficult or impossible to model (e.g., memory/cache hierarchy, addressing modes, kernel optimization). Rather than trying to model each of these factors for all kinds of hardware, we choose to define \textbf{algorithmic} compute requirements, which are independent from particular choices of hardware:

\textbf{Algorithmic FLOPs} are the number of FLOPs required to perform the mathematical calculation of a compute graph op (note: either floating point or integer arithmetic). For example, algorithmic FLOPs include the multiplies and accumulations in a matrix multiply op. Algorithmic FLOPs do not include other instructions executed by hardware to perform the computation, such as address, loop invariant, or branch target calculations. Hardware instructions that are not counted in algorithmic FLOPs are likely to account for at most constant overhead per algorithmic FLOP.

Unlike more general applications, DL compute graphs also perform backward propagation (``backprop'') of gradients from the model's predictions. Ops in a DL compute graph are differentiable, so that the gradient of each input can be calculated when given gradients of the outputs. After backprop, accumulated gradients are used to update weights and improve the model's predictions. A compute graph's backprop has highly analogous ops to the forward graph traversal, but it splits gradients to flow to model weights and to activations. The backprop for matrix operations usually has twice the algorithmic FLOPs as the forward traversal.

Analogously, we define an op's \textbf{algorithmic bytes accessed} as the total memory bytes that an op must read as inputs and write as outputs to perform the operation. Algorithmic op bytes do not include intermediate data or other memory that might be used to perform the operations, and ignores hardware effects such as caching.

We also define \textbf{algorithmic memory footprint} as the minimum number of memory bytes that must be allocated to execute a training step. More precisely, it is the minimum---over all correct topological compute graph traversals---of the maximum memory capacity required to accommodate all active tensors during any step of the traversal. Active tensors are those produced by an op in a previous traversal step, but not yet consumed by each of its downstream ops.

Finally, \textbf{algorithmic IO} counts the amount of data accessed for input to and output from a model. Training data
is often stored on disks, read from the disk, and placed into the model's input memory allocations. Algorithmic IO is
proportional to the batch size, but stays fixed as model size and training step compute requirements grow. We do not
investigate algorithmic IO further in this work, because we
expect IO will grow very slowly relative to compute.

%
\subsection{Image Classification}

\begin{figure}[bht]
  \centering
  \begin{subfigure}[bht]{\columnwidth}
    \includegraphics[width=\textwidth]{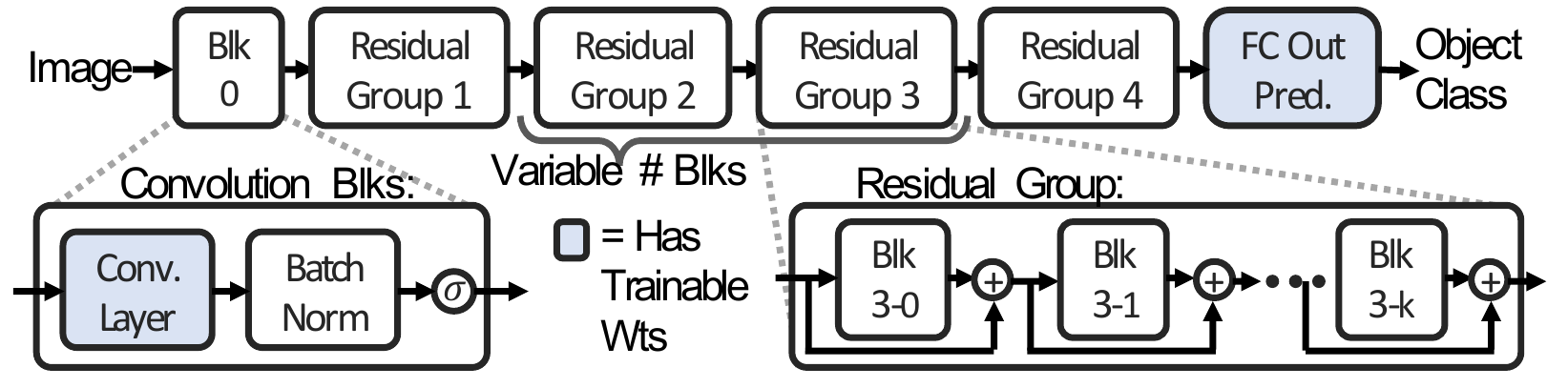}
  \end{subfigure}
  \caption{ResNet Bottleneck Abstract Architecture.}
  \label{fig:resnet_diagram}
\end{figure}

ResNets are recognized as high-accuracy convolutional networks (CNNs) for image classification and processing~\cite{he:resnets:cvpr:2016}. The ResNet bottleneck architecture, diagrammed in Figure~\ref{fig:resnet_diagram}, shows the generic structure of residual groups that allow the model grow in depth. Each residual group contains blocks of layers, as well as skip connections that permit activations to bypass the blocks. Blocks contain convolutions (with trainable weights designated in blue), batch normalization, and nonlinearities. The final layer is a fully-connected (FC) projection that maps its input to the object classes. These networks tend to be compute intensive due to their depth (50+ convolutions with 64--2048 filters each). However, as we show later, the following recurrent networks can also require significant compute.

%
\subsection{Language Modeling}

Language models (LMs) predict the next word or character given a previous sequence of input text. Most computation in these RNNs occurs in recurrent layer matrix multiplies.

\noindent\textbf{Word Language Models:}

\begin{figure}[bht]
  \centering
  \begin{subfigure}[bht]{\columnwidth}
    \includegraphics[width=\textwidth]{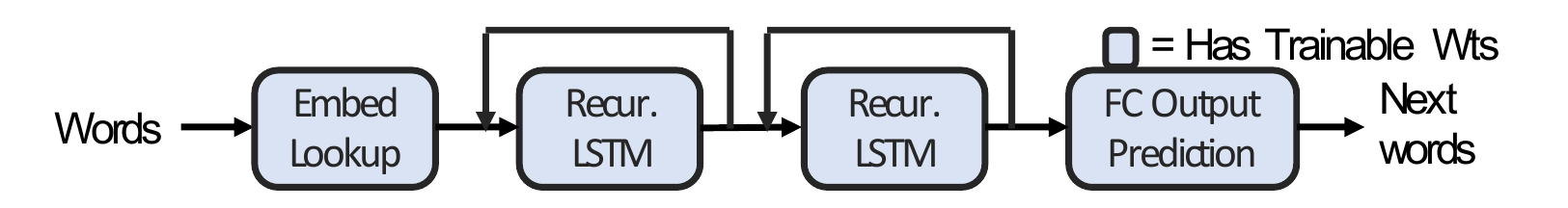}
  \end{subfigure}
  \caption{Word LM Recurrent LSTM Abstract Architecture.}
  \label{fig:word_lm_diagram}
\end{figure}

LSTM-based RNNs are the SOTA architecture for word LMs \cite{jozefowicz:lmlimits:arxiv:2016}. Figure~\ref{fig:word_lm_diagram} shows the word LM LSTM generic architecture: an embedding layer followed by recurrent layers that feed a FC output layer. The embedding layer is a table lookup operation with no algorithmic FLOPs, but it accounts for a large portion of overall weight memory footprint. The LSTM layers are moderately compute-intensive due to their many matrix multiplications in separate recurrent steps. Finally the FC output layer is compute-intensive and responsible for a large portion of activation memory footprint.

\noindent\textbf{Character Language Models:}

\begin{figure}[bht]
  \centering
  \begin{subfigure}[bht]{\columnwidth}
    \includegraphics[width=\textwidth]{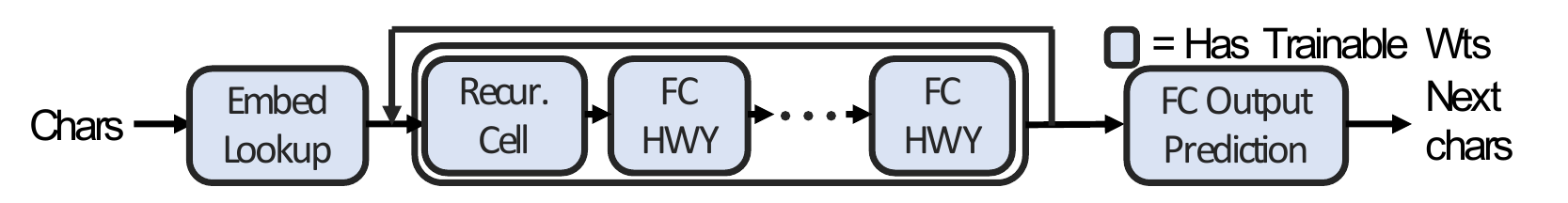}
  \end{subfigure}
  \caption{Character LM RHN Abstract Architecture.}
  \label{fig:char_lm_diagram}
\end{figure}

Recurrent-highway networks (RHN) have been shown to provide low character perplexities for character LMs \cite{zilly:rhns:icml:2017}. This character LM architecture, as depicted in Figure~\ref{fig:char_lm_diagram}, is a sequence of embedding layer, followed by a deep RHN layer, followed by an output layer. Unlike word LMs, embedding layer and output layer account for a small portion of run time and memory footprint as the vocabulary size (number of characters) is significantly smaller. Each RHN layer contains a sequence of feed-forward sublayers, and the last sublayer output feeds into the next time-step. These layers tend to be compute-intensive and responsible for a large portion of activation memory footprint, especially given their many recurrent steps (100--300 per sample).

%
\subsection{Neural Machine Translation (NMT)}

\begin{figure}[bht]
  \centering
  \begin{subfigure}[bht]{\columnwidth}
    \includegraphics[width=\textwidth]{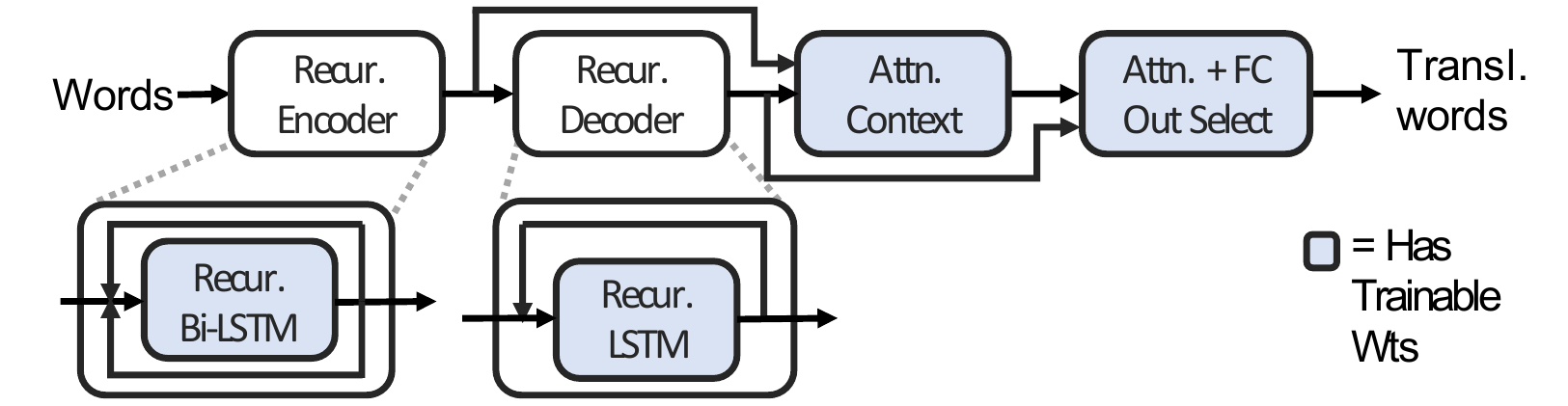}
  \end{subfigure}
  \caption{NMT Attention Abstract Architecture.}
  \label{fig:nmt_attn_diagram}
\end{figure}

The SOTA models for neural machine translation use encoder-decoder architectures with an attention mechanism to identify important recent time steps \cite{luong:globalattention:emnlp:2015}. Figure~\ref{fig:nmt_attn_diagram} diagrams such an architecture, which uses a recurrent bi-directional LSTM in the encoder and a standard LSTM for the decoder cell. The encoder and decoder feed the attention context and selection layers that choose the best decoder outputs to predict translated words. Most compute and memory access comes from the recurrent cells in this model.

%
\subsection{Speech Recognition}

\begin{figure}[bht]
  \centering
  \begin{subfigure}[bht]{\columnwidth}
    \includegraphics[width=\textwidth]{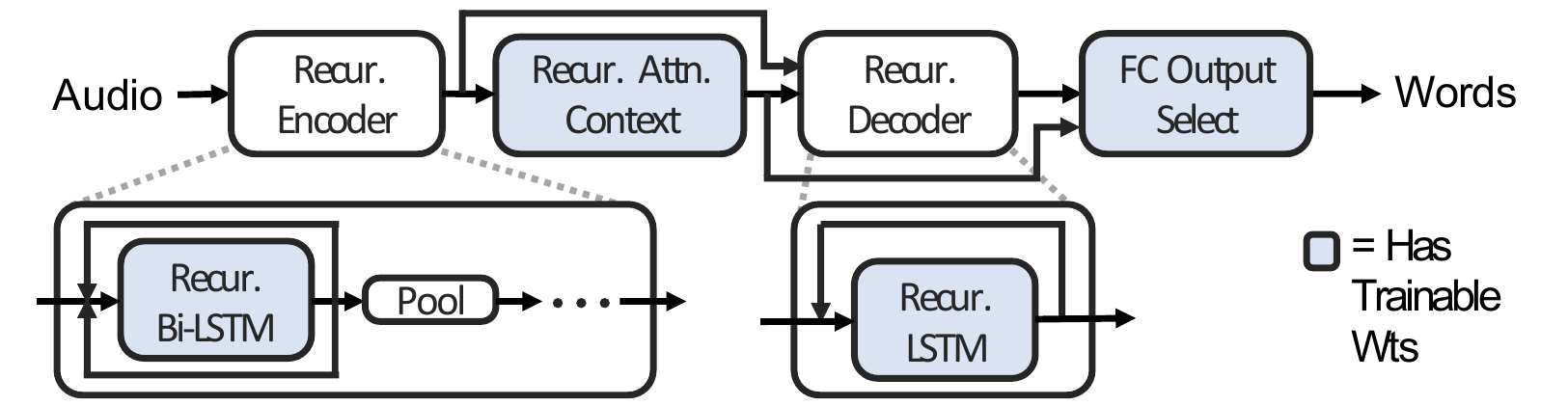}
  \end{subfigure}
  \caption{Speech Attention Abstract Architecture.}
  \label{fig:speech_attn_diagram}
\end{figure}

We investigate the hybrid attention speech model diagrammed in Figure~\ref{fig:speech_attn_diagram}~\cite{battenberg:speechtxducers:asru:2017}. Like NMT, this model is an encoder-decoder model with attention, though the encoder is a multi-layer bi-directional LSTM with intermediate pooling layers. Most computation occurs in these encoder layers. This model contains convolutions in its attention context layer, but they are very small relative to recurrent portions of network.

%% file: app_scaling.tex
\section{Application Accuracy Scaling}
\label{sec:app_scaling}

The DL community has progressively increased dataset and model sizes, and the future system demands of DL training will continue to grow. We would like to project the computational requirements of future DL applications based on the way we expect applications to grow. Recent prior work allows us to project application-level characteristics using analytical models that show the relationships between DL dataset size, model size, and model accuracy. We collect desirable accuracy targets and use the analytical models to predict the dataset and model sizes required to achieve the target accuracy. Compared to current SOTA, DL domains would like $33$--$971\times$ as much data and $6.6$--$456\times$ larger models!

%
\subsection{Motivation to Grow Data and Models}

The DL community has continually grown datasets, with open-source sets larger than 10s of GBs, to increase modeling task difficulty and model accuracy. Industry is already using significantly larger datasets. Google Research recently showed the importance of training image classifiers with $300\times$ more images than prior datasets~\cite{sun:dataeffective:iccv:2017}. Baidu's prior work uses speech recognition datasets of multiple terabytes \cite{hannun:deepspeech:arxiv:2014}. Google has also stated they want to train language models on a trillion word dataset~\cite{shazeer:mixtoreofexperts:arxiv:2017}. Such datasets of interest to DL industry are upwards of 5TB, or about 50$\times$+ larger than current publicly available datasets.

As datasets grow, DL models must also grow to fit the larger datasets, and industry is aiming for very large models. Google states they would like to train a trillion parameter model on a trillion word dataset~\cite{shazeer:mixtoreofexperts:arxiv:2017}. That same work proposes a method to compose numerous small models to reach the trillion parameter size. Our projections indicate models will easily reach into the 100s of billions of parameters. Such models would be $10$--$500\times$ larger than DL models described in current research.

%
\subsection{Accuracy Scaling with Training Data Growth}

Recent work indicates why industry wants to increase dataset and model sizes. Hestness et al. show that on real datasets, DL model accuracy improves predictably with training dataset size \cite{hestness:dlpredictscaling:arxiv:2017}. They further show that the model size required to fit the data grows predictably with data size. Industry can use these empirical models to estimate the amount of training data and model sizes required to achieve particular accuracy.

\begin{figure}[bht]
  \centering
  \begin{subfigure}[bht]{0.7\columnwidth}
    \includegraphics[width=\textwidth]{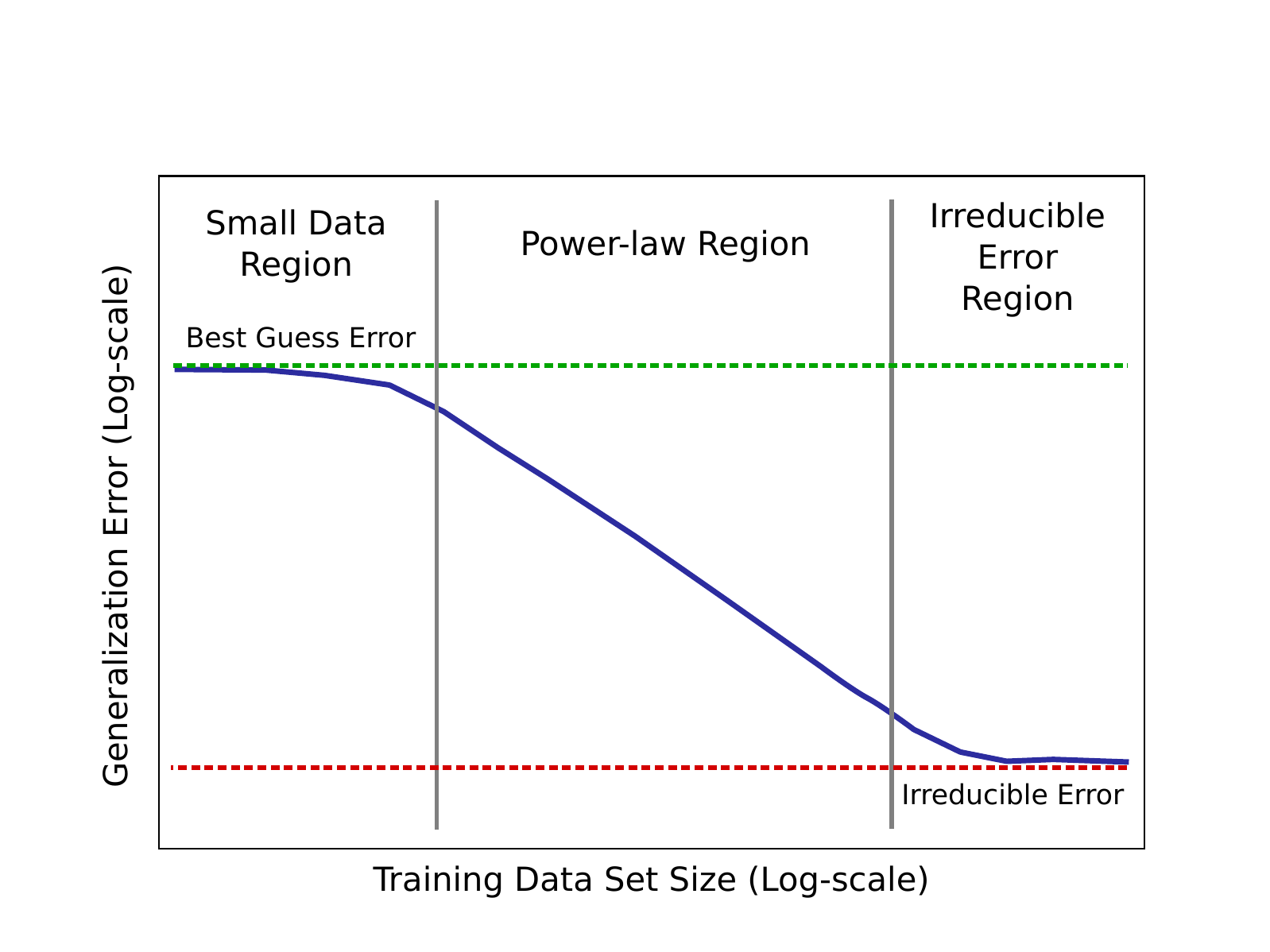}
  \end{subfigure}
  \caption{Sketch of power-law learning curves.}
  \label{fig:cartoon_power_law}
\end{figure}

Figure~\ref{fig:cartoon_power_law} shows a sketch of a model's learning curve---the reduction in prediction error as datasets grow\footnote{Copied with author permission~\cite{hestness:dlpredictscaling:arxiv:2017}}. 
The curve begins in the \textbf{small data region}, where models can only perform as well as "best" guessing for the output data distribution. The \textbf{power-law region} is where each new training sample offers information to help models improve predictions on previously unseen samples. Error declines predictably. Finally, for real applications, curves are likely to end in an \textbf{irreducible region} where models cannot further improve due to the stochastic nature of the data.

\vspace{-12pt}

In particular, we project learning curves starting from the power-law region, where most existing large-scale data applications are currently. In this region, model generalization error scales roughly as a power law:
\begin{equation}
    \varepsilon(m) \approx \alpha m^{\beta_g}
\end{equation}
Here, $m$ is the number of samples in the training dataset, and $\alpha$ and $\beta_g \in [-0.5, 0]$ are constants that depend on the structure of the modeling task, possibly including the data distribution and model architecture. $\alpha$ represents aspects of the input data space and the DL model architecture. $\beta_g$ is the power-law exponent and indicates the difficulty for models to learn more information from each additional training  example. $\beta_g$ closer to $-0.5$ means models can learn quickly from smaller datasets. Table~\ref{table:dl_domain_constants} lists estimates of $\alpha$ and $\beta_g$ for the different modeling tasks as found in the prior work ~\cite{hestness:dlpredictscaling:arxiv:2017}.

\begin{table*}[bht]
    \centering
    \small
    \caption{Learning Curve and Model Size Scaling Relationships for DL Domains}
    \begin{tabular}{|l||r|r|r|r||r|r||r|r||r|r|}
        \hline
        & & \multicolumn{1}{c|}{Desired} & \multicolumn{2}{c||}{Current Data Size} & \multicolumn{2}{c||}{Learn Curve} & \multicolumn{2}{c||}{Model Size} & \multicolumn{2}{c|}{Projected Scale} \\
        Domain (model) & \multicolumn{1}{c|}{Current SOTA} & \multicolumn{1}{c|}{SOTA} & \multicolumn{1}{c|}{Samples} & \multicolumn{1}{c||}{GB} & \multicolumn{1}{c|}{$\alpha$} & \multicolumn{1}{c||}{$\beta_g$} & \multicolumn{1}{c|}{$\sigma$} & \multicolumn{1}{c||}{$\beta_p$} & \multicolumn{1}{c|}{Data}  & \multicolumn{1}{c|}{Model} \\
        \hline
        \hline
        Word LMs (LSTM)                   & $3.37$ nat/word &   $2.48$~\cite{shannon:englishtextentropy:1951} &   $768M$ word & $3.9$ & $13.0$ & $-0.066$ & $9.4\mathrm{e}{-4}$ & $0.68$ & $100\times$ & $23\times$ \\
        Character LMs (RHN)               & $1.30$ bit/char &   $0.70$~\cite{shannon:englishtextentropy:1951} & $3.48B$ char. & $3.9$ & $9.39$ & $-0.092$ & $1.2\mathrm{e}{-5}$ & $0.89$ & $\bf 971\times$ & $\bf 456\times$ \\
        NMT (enc/dec+attn)                &     $28\%$ WPER &   $12\%$ &     $130M$ WP & $2.6$ & $3.06$ & $-0.128$ & $6.4\mathrm{e}{-4}$ & $0.68$ & $\bf 750\times$ & $90\times$ \\
        Speech Recogn. (enc/dec+attn) &     $9.5\%$ CER &    $4\%$~\cite{xiong:conver-asr-human:microsoft:2017} &  $425M$ char. &  $1674$ & $30.5$ & $-0.291$ & $2.4\mathrm{e}{-3}$ & $0.54$ &  $33\times$ & $6.6\times$ \\
        Image Classification (ResNet) &  $19.4\%$ Top-1 &    $5\%$~\cite{russakovsky:imagenet:arxiv:2015} &  $1.3M$ image &  $152$ & $15.0$ & $-0.309$ & $2.0\mathrm{e}{-2}$ & $0.57$ &  $81\times$ & $12\times$ \\
        \hline
    \end{tabular}
    \label{table:dl_domain_constants}
    \vspace{-6pt}
\end{table*}

To extend the work of Hestness et al. to predict the required data and model size from these models, we need to define accuracy targets that would be desirable for DL-enabled products. We collect feedback from DL experts and refer to prior studies that estimate the irreducible error to select desirable accuracy targets for each domain. For example, word and character LM desired SOTA are near estimated lower bounds on the entropy of English text \cite{shannon:englishtextentropy:1951}. The ``Desired SOTA'' column of Table~\ref{table:dl_domain_constants} reflects these projections.

Finally, given these analytical learning curves and target error rates, we solve the analytical models for the required data size to realize the target. The ``Projected Scale'' columns in Table~\ref{table:dl_domain_constants} show the relative data size projections. Desired SOTA values are $1.4\times$ to $3.9\times$ better than current SOTA values. However, the amount of data required to achieve these values range from $33\times$ more for speech recognition to $971\times$ more for character LMs. Language domains require the most data due to their poorer power-law exponents, $\beta_g$.

%
\subsection{Model Size Scaling with Training Data Growth}

As datasets grow in size, models must also grow in size to represent the data. Hestness et al. also collect and characterize model sizes required to fit varying training set sizes. Model parameters (roughly capacity) are expected to grow sublinearly in the training set size with the following form:\vspace{-6pt}
\begin{equation}
    p(m) \approx \sigma m^{\beta_p}\vspace{-5pt}
\end{equation}
Here, $m$ is the the number of samples in the training set, and $\sigma$ and $\beta_p \in [0.5, 1)$ are constants that depend on the problem structure, possibly including data distribution and model architecture. Models should grow parameter count more slowly than the training set (i.e., $\beta_p \leq 1$), or we could just store the dataset rather than training a model. Recent prior work shows that deep neural network model capacity---the volume of concepts (data) it can learn---grows with $O(lp\log{p})$, where $l$ is a measure of the model's depth~\cite{bartlett:dnnvcdim:colt:2017}. Loosening this bound slightly, model size should grow at least with a square root of the dataset size (i.e., $\beta_p \geq 0.5$).

Table~\ref{table:dl_domain_constants} shows empirically collected $\sigma$ and $\beta_p$ for the DL domains~\cite{hestness:dlpredictscaling:arxiv:2017}. Given the target data size determined in the last subsection, we project the model sizes required to fit the target dataset sizes. The model scale column shows the relative required increase in model size. For example, current SOTA word LMs use roughly $1B$ parameters to fit roughly $1B$ word datasets. Thus, to fit a $100\times$ larger dataset, a model would require \textasciitilde$23B$ parameters ($23$--$92$GB, depending on weight precision).

%% file: compute_scale.tex
\section{Characterizing Compute Requirements}
\label{sec:compute_scale}

Now that we have an idea of desirable data and model sizes, we turn our attention to characterizing the computational requirements to train these very large models. This section characterizes DL application compute FLOP, memory access, and memory footprint growth. Although the structure of DL applications is intricate, their training requirements scale mostly predictably. Compute and memory usage grow asymptotically linearly with model size and batch size. We provide these accessible first-order models of compute requirements not characterized in prior work.

%
\subsection{Methodology}

We estimate model training compute requirements by collecting statistics from training runs and assembling analytical models to project growth. We train with Tensorflow 1.5.0~\cite{tensorflow:whitepaper:2015} running on NVIDIA GPUs and using a modified version of TFprof. TFprof annotates compute graph ops to calculate their algorithmic FLOPs and bytes, and collect run time as they execute. At the end of a training step (i.e., a compute graph traversal), TFprof returns this profile for all ops executed during the step, ensuring that we profile even fine details of an end-to-end training step. We also query Tensorflow's memory allocators for the maximum amount of training step memory allocated---the memory footprint.

We collect profiles from 100-500 randomly-chosen training steps to account for per-training-step differences in FLOPs and memory accessed for different models. For instance, character LMs, NMT, and speech models unroll their recurrent layers for the time-steps required for the longest batch sample. This unrolling results in variable computation and memory access in separate training steps, so we average the profiled results over the training steps.

The most complicated variable to control for is training batch size---the number of data parallel samples to observe in a single training step. Batch size can be set arbitrarily, but particular batch sizes result in best model accuracy depending on data set size~\cite{smith:bayesiangeneralize:arxiv:2017}. For tested domains in this study, SOTA models have been trained using data parallelism across GPUs to increase batch size beyond the maximum memory capacity of a single GPU. It is likely that future DL training will also be constrained by per-compute-unit memory capacity, suggesting that ML researchers will choose per-compute-unit batch sizes (henceforth, ``\textbf{subbatch size}'') that can provide near-peak utilization of compute unit resources. We profile with the smallest such subbatch size.

To grow models, we change hyperparameters that have the largest effect on the ability for the model to fit larger data sets as measured by generalization error. For ResNets, increasing depth and convolution channels, rather than filter sizes, improves accuracy the most, so we collect profiles for deeper and wider image classification networks. Most recurrent models have already grown to a depth such that increased depth results in no accuracy improvement. Instead, we increase the number of hidden weights per layer.

Finally, we aim to project forward the compute requirements for models as we scale up data set and model size. The analytical models of application characteristics below use first-order approximations to provide the community with a concise set of formulas for projections. However, we also use high-fidelity modeling to verify these results (Appendix~\ref{sec:appendix_artifact}).

%
\subsection{Estimating Training Step Algorithmic FLOPs}

For DL models, the number of FLOPs per training step grows roughly linearly in the number of parameters of the model, suggesting that each model parameter is used roughly the same number of times in a single training step. We demonstrate this observation analytically for word LMs next.

Again, let $p$ be the number of model parameters for a LSTM word LM, and let $p_{em}$, $p_{re}$, and $p_{o}$ be the parameters in embedding, recurrent, and output layers, respectively. We approximate the total model parameters as:
\begin{equation*}
    p = p_{em} + p_{re} + p_{o} \approx hv + 8h^2l + hv = 8h^2l + 2hv
\end{equation*}
Here, $v$ is the LM's vocabulary size, $h$ is number of hidden weights per recurrent layer, and $l$ is the number of layers.

Next, we show the roughly linear relationship between parameters and FLOPs per step. Since backward propagation adds \textasciitilde$2\times$ the number of FLOPs, regardless of the model, we consider only the forward propagation. For this first-order model, we assume that most compute FLOPs come from the subset of ops that perform vector or matrix operations. We estimate forward propagation algorithmic FLOPs:
\begin{equation*}
    c_{fwd} = c_{em} + c_{re} + c_{o} \approx 0 + 16lqh^2 + 2qvh = q(16h^2l + 2hv)
\end{equation*}
Here, $q$ is the sequence length for the training step (we ignore subbatch size to normalize per training sample). These models indicate that $\lim_{h\rightarrow\infty} \frac{c_{fwd}}{p} \rightarrow k$, a constant. Thus, we expect that for word LMs and similarly structured recurrent models, compute FLOPs should grow roughly linearly in the increase in number of model parameters.

\begin{figure}[bht]
  \centering
  \begin{subfigure}[bht]{\columnwidth}
    \includegraphics[width=\textwidth]{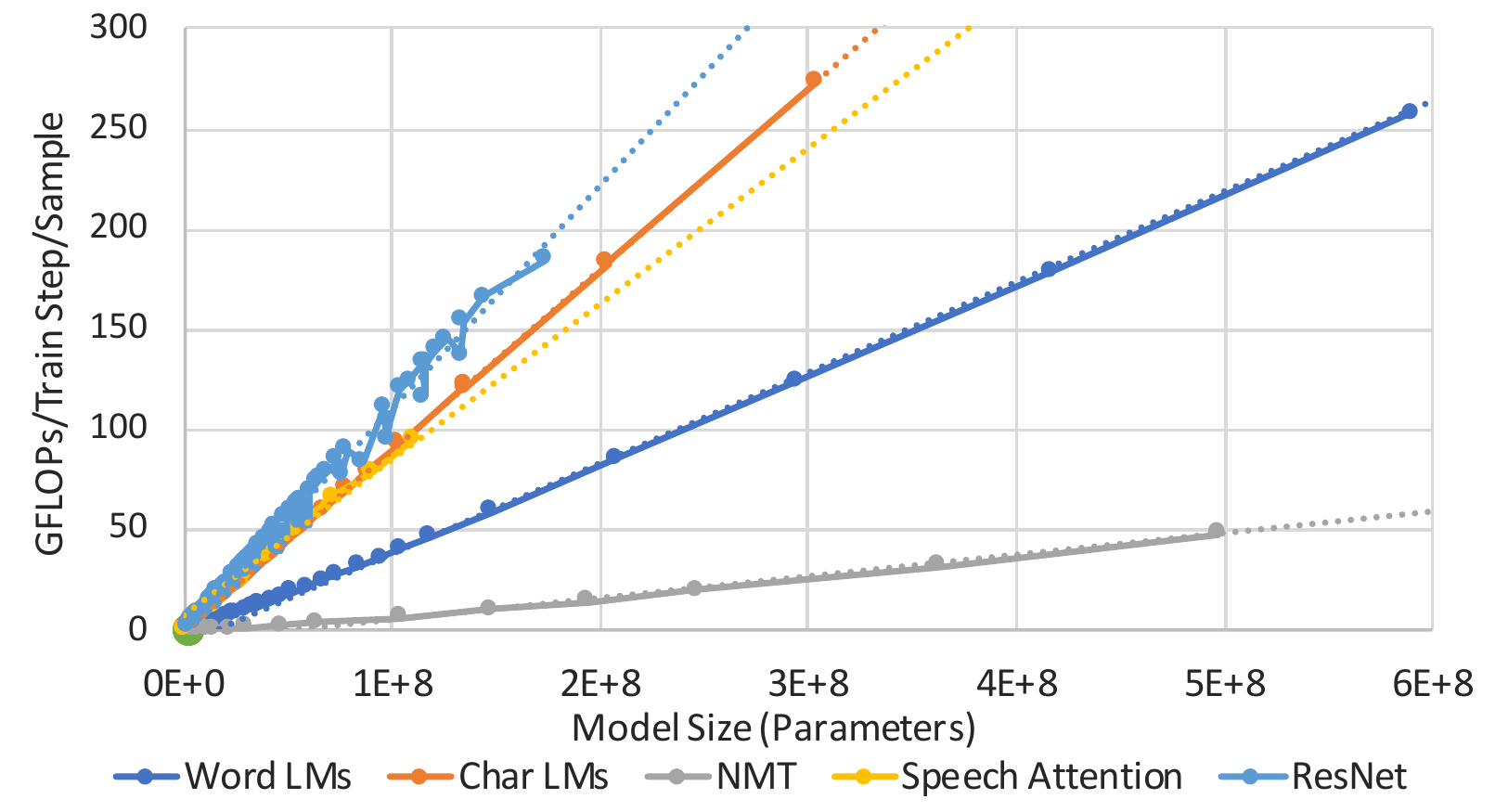}
  \end{subfigure}
  \caption{Per-training sample FLOPs growth with number of model parameters, all models (dotted lines are trends).}
  \label{fig:flops_projections}
\end{figure}

We confirm this linear relationship between model parameters and algorithmic FLOPs per training step empirically across our set of applications. Figure~\ref{fig:flops_projections} plots the TFprof-profiled growth in algorithmic FLOPs (note: batched training roughly multiplies these values by the subbatch size). Each domain's algorithmic FLOPs grow linearly with model size above $30$--$100M$ parameters---moderately large models. FLOPs per parameter ranges from 149 for NMT to 1111 for ResNets. For recurrent networks, as sequence length grows, the FLOPs/parameter also grows, approaching ResNet requirements. Character LMs and speech networks unroll layers for 150 and 300 time-steps, respectively.

\vspace{6pt}

Table~\ref{table:dl_domain_requirements} records the asymptotic hardware requirements for each DL domain as models grow. Given the clear linear relationships between FLOPs and parameter counts, we use the following linear trend to project the compute FLOPs per training sample ("$c_t$") for models with $p$ parameters:
\begin{equation*}
    c_t(p) \approx \gamma p
\end{equation*}
Here, $\gamma$ is a constant that depends on the input data shape, recurrent sequence length, and model architecture.


\begin{table*}[bht]
    \centering
    \small
    \caption{Asymptotic Application-level Compute Requirements}
    \begin{tabular}{|l||r|r|r|r|}
        \hline
        & \multicolumn{1}{c|}{Alg. compute} & \multicolumn{1}{c|}{Alg. memory access} & \multicolumn{1}{c|}{Alg. operational intensity} & \multicolumn{1}{c|}{Minimal Mem. Foot} \\
        Domain (model) & \multicolumn{1}{c|}{(FLOPs/Param)} & \multicolumn{1}{c|}{(Bytes/Param)} & \multicolumn{1}{c|}{(FLOPs/Byte)} & \multicolumn{1}{c|}{(Bytes/Param)} \\
        \hline
        \hline
        Word LMs (LSTM)               & $481~b$ &  $1755+30784~b/\sqrt{p}$ &  $b\sqrt{p}/(3.65\sqrt{p} + 64~b)$ & $11.94$ \\
        Character LMs (RHN)           & $900~b$ & $3510+102980~b/\sqrt{p}$ &  $b\sqrt{p}/(3.9\sqrt{p} + 114~b)$ & $12.47$ \\
        NMT (enc/dec+attn)            & $149~b$ & $533+22653~b/\sqrt{p}$ & $b\sqrt{p}/(3.6\sqrt{p} + 151~b)$ & $10.32$ \\
        Speech Recogn. (enc/dec+attn) & $775~b$ & $3100+162750~b/\sqrt{p}$ &  $b\sqrt{p}/(4.0\sqrt{p} + 210~b)$ & $32.94$ \\
        Image Classification (ResNet) & $1111~b$ & $66.7+268862~b/\sqrt{p}$ & $b\sqrt{p}/(0.06\sqrt{p} + 242~b)$ & $42.57$ \\
        \hline
    \end{tabular}
    \label{table:dl_domain_requirements}
    \vspace{-10pt}
\end{table*}

%
\subsection{Estimating Algorithmic Memory Bytes Accessed\label{subsec:alg_mem_access}}

Like algorithmic FLOP counts, algorithmic memory accesses also scale linearly with model parameters across the DL applications. However, since a significant portion of training step memory accesses are from reading or updating model weights---which do not depend on the subbatch size---memory access counts depend, to first-order, on both model size and subbatch size. This section describes an analytical model and verifies that it fits empirical results.

\begin{figure}[bht]
  \centering
  \begin{subfigure}[bht]{\columnwidth}
    \includegraphics[width=\textwidth]{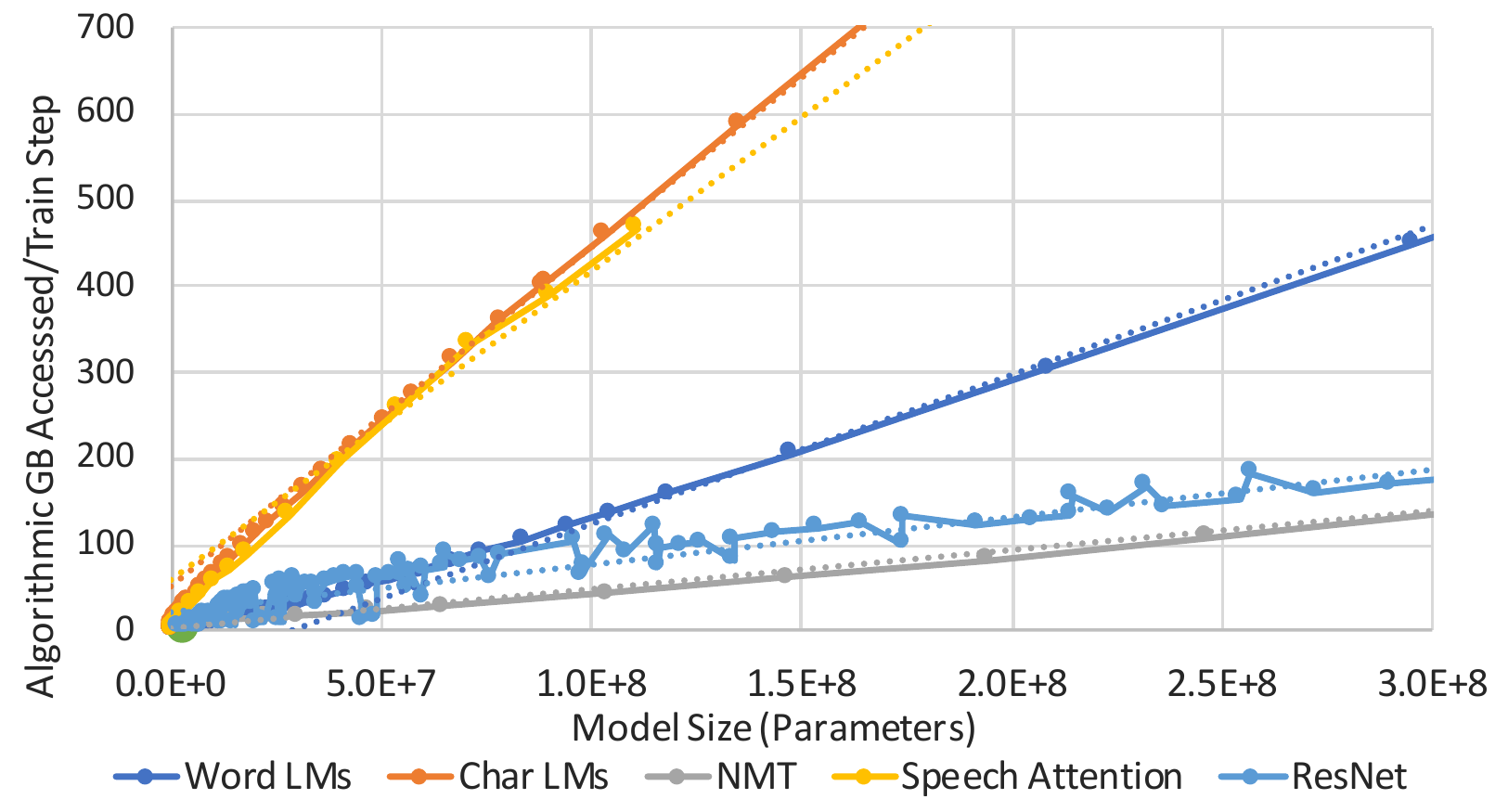}
  \end{subfigure}
  \caption{TFprof algorithmic memory access as model size grows (note: for particular subbatch sizes).}
  \label{fig:bytes_growth_projections}
\end{figure}

A training step must access two types of tensors: the DL model and the activation tensors that flow through the model. Hardware loads from and stores to the model parameters roughly a constant number of times each for the forward and backward propagation, and to update the weights at the end of a training step. Similarly, activation memory, with dimensions proportional to the batch size and model dimensions, is accessed roughly a constant number of times. As above, denote $s$ as the model parameter count. Then total memory accesses for a training step ("$a_t$") takes this first-order form\footnote{Supplemental material shows detailed calculation for word LMs}:
\begin{equation*}
    a_t(p, b) \approx \lambda p + \mu b \sqrt{p}
\end{equation*}
Here, $\lambda$ and $\mu$ are constants that depend on input data shape, recurrent sequence length, and model architecture. The $\sqrt{p}$ term approximates the model’s hidden layer weight or channel counts---one dimension of the compute graph's large linear algebra ops. We find $\sqrt{p}$ is a good approximation for all domains, with a small caveat: For models with many parameters to embed input data (e.g., the larger vocabularies of word LMs and NMT), $\sqrt{p}$ over-estimates hidden dimension until the hidden dimension is large relative to the embedding dimension. Figure~\ref{fig:bytes_growth_projections} curves show nearly linear asymptotes.

\vspace{6pt}

%
\subsection{Estimating Training Operational Intensity}

Conveniently, although model training steps are composed of many ops, their algorithmic FLOPs and memory access characteristics are strikingly similar to those of a single large linear algebra operation. As a result, operational intensity---the ratio of FLOPs to memory bytes accessed---takes form familiar in linear algebra kernel optimization.

Algorithmic operational intensity for each DL model is listed in Table~\ref{table:dl_domain_requirements}. A model's ops that contribute the most to FLOPs and memory accesses are often matrix operations with dimensions related to the hidden dimension (\textasciitilde$\sqrt{p}$) and subbatch size. The operational intensity of a matrix multiplication with dimensions ($b\times\sqrt{p}$)($\sqrt{p}\times\sqrt{p}$) is $b\sqrt{p}/(2\sqrt{p} + 4~b)$, the same form as the end-to-end training step operational intensities listed in Table~\ref{table:dl_domain_requirements}.

\begin{figure}[bt]
  \centering
  \begin{subfigure}[bht]{\columnwidth}
    \includegraphics[width=\textwidth]{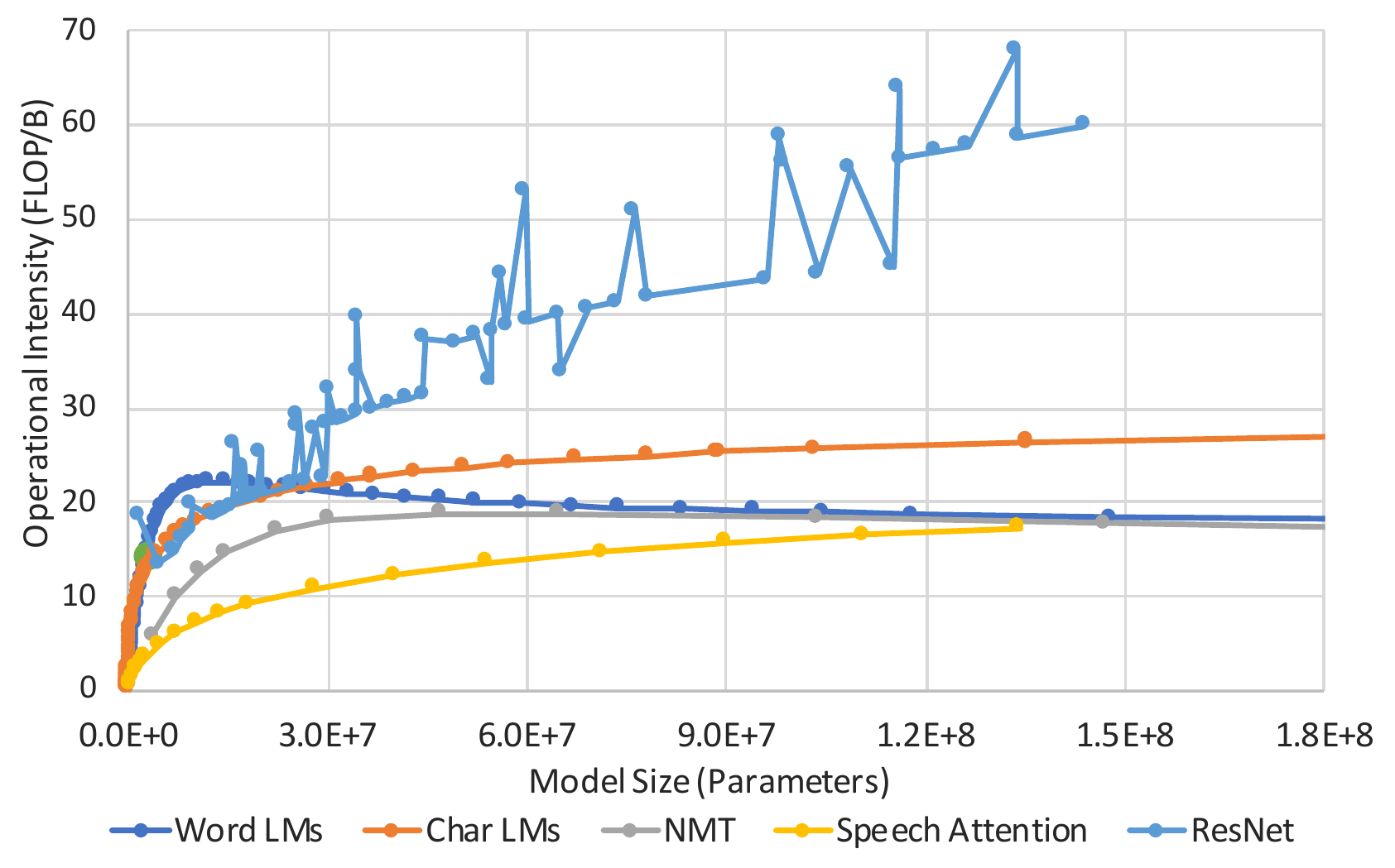}
  \end{subfigure}
  \caption{TFprof algorithmic operational intensity as model size grows (note: fixed subbatch size).}
  \label{fig:compute_intensity_projections}
\end{figure}

As a result of its form, operational intensity will approach some fixed upper bound unless both a model's hidden dimension and the subbatch size grow. When either model size or subbatch size is fixed, it will asymptotically approach the ratio of the slopes of algorithmic FLOPs and bytes growth. Figure~\ref{fig:compute_intensity_projections} shows the leveling of operational intensity for fixed subbatch size as model size grows for each application.

\vspace{3pt}

%
\subsection{Estimating Training Step Memory Footprint}

Memory footprint is the measure of the memory capacity required to execute an algorithm. Tensorflow's memory allocator provides a footprint estimate, but we also estimate minimal memory footprint by tracking it through a topological traversal of the compute graph. For each op, DL frameworks can allocate memory for the op's output tensors, and after executing the op, free the op's input tensors if all the tensor's consumer ops have executed.

\begin{figure}[bht]
  \centering
  \begin{subfigure}[bht]{\columnwidth}
    \includegraphics[width=\textwidth]{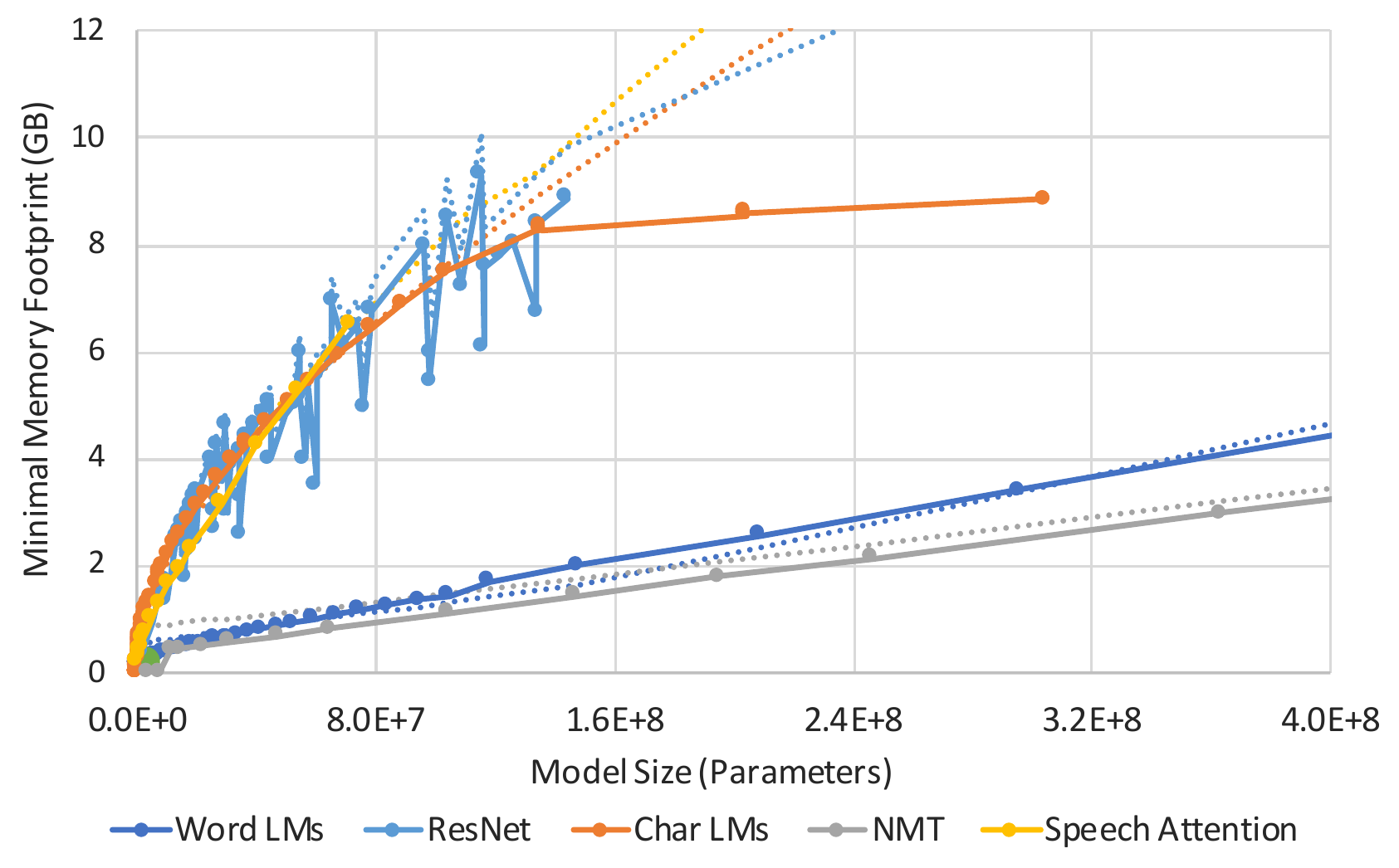}
  \end{subfigure}
  \caption{Empirical and asymptotic minimal memory footprint as model size grows (note: fixed subbatch size).}
  \label{fig:minimal_memory_footprint}
\end{figure}

Figure~\ref{fig:minimal_memory_footprint} plots the Tensorflow allocator memory footprints for each model and our topological estimates. These values agree up to the point that Tensorflow runs out of GPU memory capacity (80\% of 12GB). At that point, the allocator starts swapping GPU memory to the CPU's memory space, where it no longer counts the memory as part of the footprint. When Tensorflow does not swap memory, our models tend to slightly overestimate minimal memory footprint; Tensorflow optimizes to perform some ops on tensors in-place rather than allocating separate output tensors.

\vspace{3pt}
\vspace{2pt}

Minimal memory footprint grows asymptotically linearly with model size for larger models. This trend is expected given that the model's parameters dominate memory and are persistent, while activation tensors can be freed and reused by the framework. We model minimal footprint linearly:
\begin{equation*}
    f_t(p) \approx \delta p
\end{equation*}
Here, $\delta$ is a constant dependent on the input data shape, recurrent sequence length, and model architecture. This first-order approximation fits well for parameter counts above \textasciitilde$500M$, but for our projections in the next section, we opt to use more accurate topological traversal estimates.

Language model footprint growth is similar across the domains; character LM footprint growth slows significantly for large models (not depicted in the figure). Speech and image domains show faster memory footprint growth with model size. However, as the next section shows, speech and image domains need much smaller networks to achieve accuracy targets, so their footprint requirements are modest.

%% file: projections.tex
\section{Projecting the Accuracy Frontier\label{sec:projections}}

Here, we project the compute resources required to train models to target accuracy levels. We also project a hypothetical Roofline estimate of model training time and discuss implications of the resource requirements. Improving speech recognition and image classification should be feasible with existing parallelism strategies. Language domains, however, are likely to require $100\times$ more compute, suggesting the need for both improved algorithmic and parallelism strategies.

%
\subsection{Projecting Target Compute Requirements}

Using the analytical models from the last two sections, we project compute resource requirements to reach target accuracy levels. Table~\ref{table:target_training_requirements} lists the projected data and model size, our choice of subbatch sizes (Section~\ref{sec:subbatch_sizing}), and projected training requirements.

\begin{table*}[bht]
    \centering
    \small
    \caption{Application-level Training Requirements Projected to Target Accuracy}
    \begin{tabular}{|l||r|r|r||r|r|r|r|r|}
        \hline
        & \multicolumn{2}{c|}{Projected} & \multicolumn{1}{c||}{Sub-} & \multicolumn{3}{c|}{Training Step} & \multicolumn{2}{c|}{Accel. Time} \\
        & & \multicolumn{1}{c|}{Model} & \multicolumn{1}{c||}{batch} & \multicolumn{1}{c|}{TFLOPs/} & \multicolumn{1}{c|}{Mem Acc.} & \multicolumn{1}{c|}{Min Mem} & \multicolumn{1}{c|}{Step} & \multicolumn{1}{c|}{Epoch} \\
        Domain (model) & \multicolumn{1}{c|}{Data Size} & \multicolumn{1}{c|}{Params.} & \multicolumn{1}{c||}{Size} & \multicolumn{1}{c|}{Step} & \multicolumn{1}{c|}{TB/Step} & \multicolumn{1}{c|}{Foot (GB)} & \multicolumn{1}{c|}{(secs)} & \multicolumn{1}{c|}{(days)} \\
        \hline
        \hline
        Word LMs (LSTM)               &   $77B$ word & $23.8B$ & $128$ &  $1444$ &  $41.5$ &  $272$ &  $\bf 115$ &  $\bf 31K$ \\
        Character LMs (RHN)           & $3.4T$ char. &  $146B$ &  $96$ & $12618$ & $488.1$ & $1703$ & $\bf 1007$ & $\bf 3.5M$ \\
        NMT (enc/dec+attn)            &   $97.4B$ WP & $18.9B$ &  $96$ &   $499$ &  $18.4$ &  $185$ & $\bf 39.8$ &  $\bf 16K$ \\
        Speech Recogn. (enc/dec+attn) &  $14B$ char. &  $727M$ & $128$ &    $72$ &   $2.8$ &   $30$ &      $5.8$ &       $93$ \\
        Image Classification (ResNet) & $103M$ image &  $732M$ &  $32$ &    $28$ &   $0.4$ &   $34$ &      $2.3$ &       $84$ \\
        \hline
    \end{tabular}
    \label{table:target_training_requirements}
    \vspace{-6pt}
\end{table*}

We expect that image processing networks will require the least growth in algorithmic FLOPs and memory access per training step to achieve aggressive accuracy targets. Their required model growth is small relative to recurrent networks, and their convolutional layers offer high operational intensity to utilize compute resources with smaller subbatch sizes. The clearest contrast is with speech recognition, which would require similar model size as image classification, but its larger subbatch size means more FLOPs and memory access per training step. These results suggest it may be easier to parallelize very large image network training by sharding full batches across many accelerators.

The projected compute requirements also witness the challenges of scaling language domains specifically, and recurrent networks in general. To reach target accuracy on language and speech domains will require $2.5$--$1200\times$ more FLOPs and memory access per training step than image classification. In language domains, these increases are largely due to the model size growth required to fit larger data sets.

Finally, we note that all domains are likely to require significantly more memory capacity than available with current accelerators. Current GPUs and Google's TPU v2 have 16 or 32GB of memory per accelerator chip~\cite{dean:sysml:nips:2017}. Running any of these models on such accelerators will require either model-level parallelism to split portions of the models across multiple accelerator's memories, or migrating model parts into and out of accelerator memory---an expensive operation.

%
\subsection{Projecting Run Time on Hardware}

Next, we estimate hypothetical best-case run times for each of the target applications running on an accelerator. We configure a target accelerator, describe our process for choosing the training step subbatch size, and then estimate run time. The estimates use the Roofline model to predict the overall system throughput given the full-graph algorithmic FLOPs and memory accesses~\cite{williams:roofline:acm:2009}.

\begin{table}[bht]
    \centering
    \small
    \caption{Target Accelerator Configuration}
    \begin{tabular}{|l||r|}
        \hline
        Component & Configuration \\
        \hline
        \hline
        Compute Throughput, 32-bit ($x_c^{32b}$) & $15.67$ TFLOP/s \\
        On-chip Cache                            &          $6$ MB \\
        Memory Bandwidth ($x_a$)                 &      $898$ GB/s \\
        Memory Capacity (off-chip)               &         $32$ GB \\
        Inter-device Bandwidth                   &       $56$ GB/s \\
        \hline
    \end{tabular}
    \label{table:target_accelerator}
\end{table}
\vspace{5pt}

Table~\ref{table:target_accelerator} shows the configuration for a target accelerator similar to NVIDIA's V100 version 2. We assume maximum achievable throughput of 80\% of peak FLOPs and 70\% of peak memory bandwidth, consistent with existing hardware. The accelerator's compute intensity inflection point between memory-bound and bandwidth-bound (its Roofline ``\textbf{ridge point}'') is $17.4$ FLOP/B, but given peak achievable throughput, rises to $19.9$ FLOP/B. We start by assuming that the accelerator has infinite memory capacity and is able to fit the memory footprint for a training step of any model.

\subsubsection{Subbatch Size: Minimize Per-Sample Time\label{sec:subbatch_sizing}}

Choosing an appropriate subbatch size for model training is a difficult process that depends on many aspects of the DL application. Here, we focus on the hardware trade-offs: we want to ensure good utilization of the accelerator while keeping a small memory footprint. We identify three subbatch size points-of-interest and show that the smallest size that minimizes per-sample latency (i.e., maximizes throughput) provides the best trade-offs.

\begin{figure}[bt]
  \centering
  \begin{subfigure}[bht]{\columnwidth}
    \includegraphics[width=\textwidth]{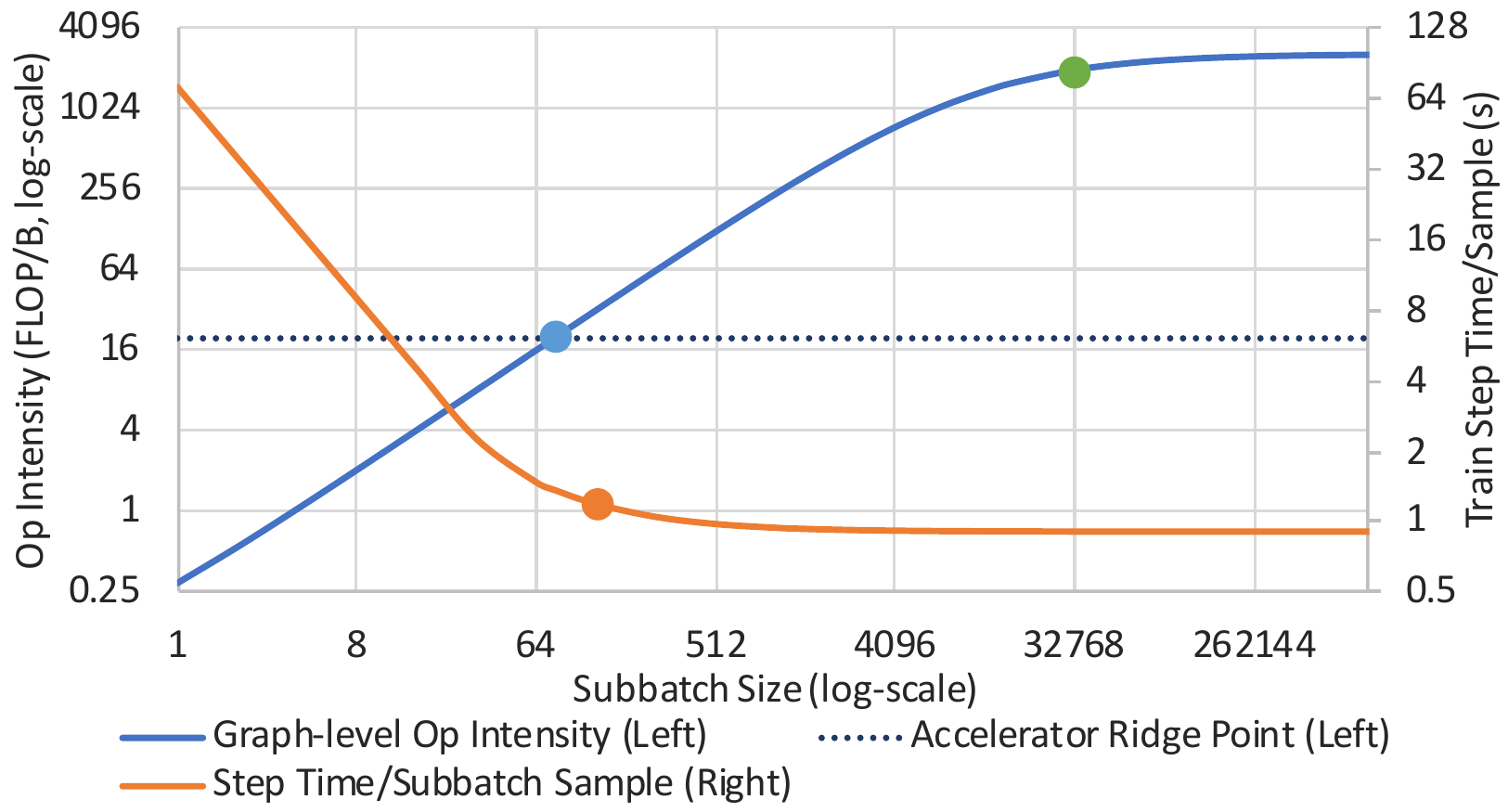}
  \end{subfigure}
  \caption{Subbatch size effect on word LM operational intensity, training step time per sample. Marginal run time gains for op intensity higher than accelerator ridge point.}
  \label{fig:subbatch_size_choice}
\end{figure}
\vspace{3pt}

Figure~\ref{fig:subbatch_size_choice} shows the effect of subbatch size on the graph-level operational intensity and the training step time per-subbatch-sample. We could choose subbatch size such that the graph-level operational intensity nears saturation (green marker), giving the most opportunity to utilize the accelerator's compute throughput. However, this point also requires a very large memory footprint, often $5$--$20\times$ more than a small subbatch. Another option is subbatch size such that the graph-level operational intensity matches the accelerator's ridge point (blue marker). In practice, however, this point does not optimize the accelerator's compute throughput---many ops are still memory-bound. The training step time-per-sample curve (orange) shows 40\% throughput loss.

Instead, we prefer subbatch size that minimizes the training step time normalized per-sample. The orange point in Figure~\ref{fig:subbatch_size_choice} is a subbatch size that keeps memory footprint small while achieving 79\% of the peak compute throughput. We use this approach to estimate best subbatch sizes for each domain in Table~\ref{table:target_training_requirements}. For recurrent networks, subbatch size settles at about $1.5\times$ larger than the point where graph-level operational intensity matches the accelerator's ridge point.

\subsubsection{Per-epoch Run Time}

Finally, we estimate best-case run time using a Roofline model---performance is bounded either by the accelerator's compute ($x_c$) or memory access ($x_a$) throughput:
\begin{equation*}
    r_t(x_c, x_a) = max\left(\frac{c_t}{80\% \cdot x_c}, \frac{a_t}{70\% \cdot x_a}\right)
\end{equation*}
We list training step time in Table~\ref{table:target_training_requirements}, and project these out to the training time for one epoch. These estimates were also used for selecting subbatch sizes.

Although optimistic, these training time projections show that the target accuracies for image classification and speech recognition may not be far out of reach. A single epoch would take \textasciitilde$3$ months on a single accelerator. Reducing epoch time to less than a day would require parallelizing training over \textasciitilde$100$ accelerators---activity becoming more common in recent data parallelism work.

However, major challenges exist in language domains with epoch times of $40$--$9600$ years on a single accelerator. To achieve target accuracies for these domains will require significant innovation beyond existing parallelization strategies.


%% file: case_study.tex
\section{Case Study: Word LMs at the Frontier}

Language and translation domains are among the most challenging problems we have tested. Our results show they may require models of \textasciitilde$20$--$200B$ parameters and \textasciitilde$100\times$ more compute than other domains. This section works through a case study of word LMs to consider the challenges and potential approaches to scale to their frontier accuracy. A combination of algorithmic and parallelism optimizations is required to train a target word LM in 7 days per epoch.

%
\subsection{Setting the Baseline Train Time}

\begin{table*}[bht]
    \centering
    \small
    \caption{Step-by-Step Process of Training Word LM to Target Accuracy.}
    \begin{tabular}{|l||r|r||r|r|r|r|}
        \hline
                           & \multicolumn{1}{l|}{Num.} & \multicolumn{1}{l||}{Batch} &   \multicolumn{1}{l|}{Accel. Mem.} &      \multicolumn{1}{l|}{L2 Cache} & \multicolumn{1}{l|}{Train Time} & \multicolumn{1}{l|}{Alg. FLOP} \\ 
        Optimization Stage & \multicolumn{1}{l|}{Accel.} &  \multicolumn{1}{l||}{Size} & \multicolumn{1}{l|}{Required (GB)} & \multicolumn{1}{l|}{Capacity} & \multicolumn{1}{l|}{days/epoch} & \multicolumn{1}{l|}{Utilization} \\
        \hline
        \hline
        Best-case (Roofline) Baseline   &    $1$ &    $128$ & \textbf{\color{darkred} 113.8} & \textbf{\color{darkred}---} &  2707 &  80\% \\ 
        Cache-hierarchy-aware Baseline &    $1$ &    $128$ & \textbf{\color{darkred} 113.8} &       6MB &  $4071$ &  $46\%$ \\
        w/ Data Parallelism (Option 1)  & $1024$ & $131072$ & \textbf{\color{darkred} 125.7} &       6MB &   $6.2$ &  $34\%$ \\
        w/ Data Parallelism (Option 2)  &  $512$ &  $65536$ & \textbf{\color{darkred} 125.7} &       6MB &  $11.1$ &  $38\%$ \\
        + Layer Parallelism ($4\times$) & $2048$ &  $65536$ &  \{\textbf{\color{darkred} 60}, 17, 17, 32\} &       6MB &  $7.2$ & $14.5\%$ \\
        + Shard the Embedding Layer     & $2048$ &  $65536$ & \{32, 31, 31, 32\} &       6MB &  $7.2$ & $14.5\%$ \\
        \hline
    \end{tabular}
    \label{table:case_study}
    \vspace{-6pt}
\end{table*}

We begin by setting the baseline training time. Since the prior word LM would take \textasciitilde$84$ years per epoch to train, we start by choosing an algorithmic optimization used in recent word models: LSTM projection~\cite{sak:lstm:arxiv:2014}. The projected LSTM reduces the inner dimension of the last hidden layer before feeding it to the output layer. We also increase the vocabulary size to match prior work~\cite{jozefowicz:lmlimits:arxiv:2016}. These changes reduce the per-training-step FLOPs, memory accesses, reducing the roofline time by a factor of $11.7\times$ to \textasciitilde$9.89s$ on the target accelerator. Table~\ref{table:case_study} records our process of parallelizing the word LM, starting with this best-case $2707$ days per epoch.

We also make our target application more realistic by adding simple modeling for memory accesses in matrix arithmetic. Unfortunately, algorithmic memory accesses underestimate total memory accesses for ops that perform large matrix multiplications; portions of the input tensors can be stored in on-chip caches of the accelerator, but large-tensor multiplies will require re-streaming significant portions of the inputs from off-chip memory multiple times. We capture the impact of cache hierarchy on performance by modeling these extra memory accesses, assuming a common, tiled matrix multiply implementation \cite{coleman:cachetiling:sigplan:1995}. This cache-hierarchy-aware model predicts per-epoch time would take $4671$ days, reducing to 46\% algorithmic FLOP utilization. We validate that this model is still optimistic, but reduces maximum prediction error from 42\% down to 15\% on tested hardware.

%
\subsection{Step-by-Step Parallelism Strategy}

There are three major challenges to scale this word LM's training time.
First, we will need to reduce training time by $667\times$ ($4671$ days$/7$), requiring parallelism across at least this many accelerators. Second, the required memory footprint is too large to fit in a single accelerator's memory, so each data-parallel worker will need to parallelize across at least $4$ accelerators ($113.8$GB per step$/32$GB capacity per accelerator). Finally, we aim for effective use of resources, and describe a parallelism scheme that keeps accelerator algorithmic FLOPs utilization above $14.5\%$.

\subsubsection{Data Parallelism}

We first scale out using data parallelism---the process of dividing batch elements across multiple workers, and then collecting their results to update model weights. The baseline subbatch size is 128, found using the method in Section \ref{sec:subbatch_sizing}. We model a synchronous stochastic gradient descent (SGD) approach implemented using a ring-allreduce \cite{patarasuk:allreduce:jpdc:2009}.

\begin{figure}[t]
  \centering
  \begin{subfigure}[bht]{\columnwidth}
    \includegraphics[width=\textwidth]{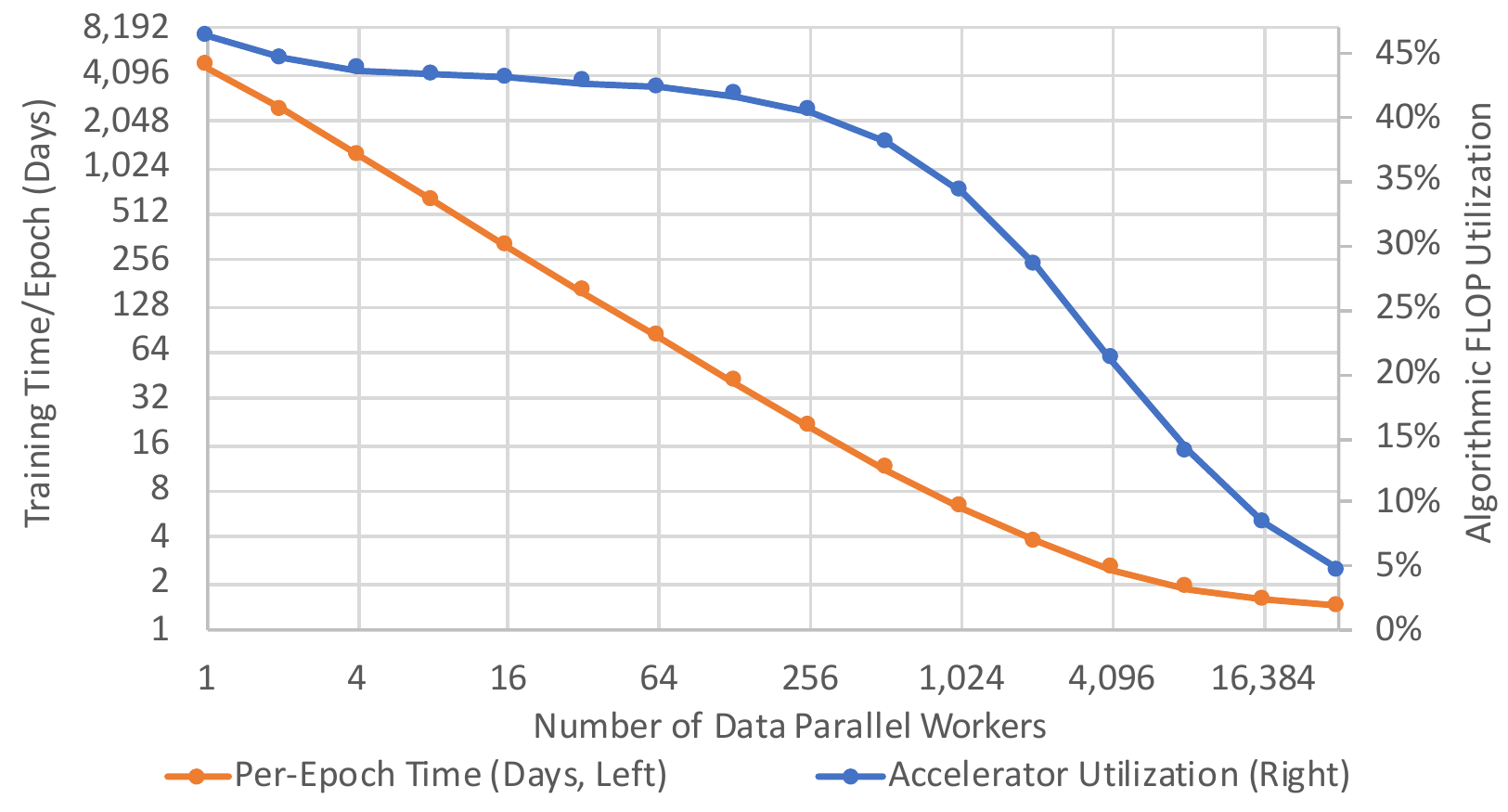}
  \end{subfigure}
  \caption{Data parallelism effect on overall run time and utilization at subbatch = 128.}
  \label{fig:data_parallel}
\end{figure}

SGD communication overheads eventually dominate per-training-step time. As listed in Table~\ref{table:target_accelerator}, we assume accelerators can communicate using high bandwidth inter-device links at $56$GB/s, consistent with future intra-node and Infiniband 400Gb inter-node interconnects. Figure~\ref{fig:data_parallel} shows training time per epoch improves while utilization declines as we increase the number of data-parallel workers.

Modeling results show that word LMs would require at least $1024$ accelerators to reduce epoch time to 6.2 days. Utilization declines slightly to 34\% at $1024$ accelerators due to communication overheads for reducing gradients. Recent prior work uses batch sizes up to $32K$ samples to train image classification~\cite{you:imagenetminutes:icpp:2018} and character LMs~\cite{puri:40gbcharmodels:arxiv:2018}, so we believe batch size $32$--$128K$ may be feasible with future techniques. We also list a configuration for $512$ accelerator data parallelism as the basis for the next stage.

\subsubsection{Model Parallelism}

Although data parallelism gets close to our desired training time, it does not address the problem of per-accelerator required memory footprint. At the current optimization stage, each accelerator would require roughly $126$GB of memory capacity, so we must divide the model to parallelize training steps across more accelerators.

We consider ``layer-wise parallelism'', an approach that places separate layers of the model across neighboring accelerators. Since the word LM has 4 layers, we allocate 4 accelerators per data-parallel worker. Starting from the $512$ data parallel worker option, we require $2048$ total accelerators to add layer parallelism. This approach would reduce epoch time to just over 7 days, and cut the required memory footprint per accelerator by half.

Although layer parallelism reduces the footprint required per accelerator, the embedding layer of the word LM ($59.5$GB) will not fit in a single accelerator's memory. Prior techniques move the embedding layer to locations with more memory capacity, such as host memory, which will require moving embedded data to the accelerator's memory. Instead, we propose to split the embedding layer into 3 pieces and locate two smaller parts in the memories of accelerators that perform recurrent layer computations. This split evens out the per-accelerator footprints with trivial run time overhead, and results in final algorithmic FLOP utilization of $14.5\%$.

\subsubsection{Discussion and Related Work}

This word LM case study highlights the challenges that will exist for scaling RNN model training to frontier-level accuracy. Major opportunities to optimize RNN training all result in the need for more cache or memory capacity.

\textbf{Memory Capacity:} Although emerging accelerators use high-bandwidth memories (HBM), their capacities are currently just $16$ or $32$GB. Existing CNN applications can utilize compute FLOPs of these accelerators with small memory footprints, so there is little pressure from these applications to increase memory capacity. On the other hand, each of the language domains show extreme per-accelerator training-step memory footprint that exceeds current memory capacities by $8$--$100\times$.

We started with algorithmic optimization to reduce the word LM's memory (and compute) requirements. Accelerator memory capacities are orders of magnitude short of LM requirements, but some algorithmic optimizations may be promising to chip away at this gap. Model compression or distillation, and low-precision or sparse computation may reduce model or activation tensor size~\cite{cheng:modelcompresssurvey:ieeesigprocmag:2018}, and reduce memory requirements by $1.5$--$10\times$. Currently, many challenges exist to use these techniques during model training.

\textbf{Parallelism Techniques:} DL frameworks could also provide parallelization techniques to enable more effective use of cache and memory capacity. Researchers have shown how to train image and language applications quickly using data-parallel scaling to reduce per-accelerator activation memory requirements \cite{goyal:largeminibatch:fb:2017, you:imagenetminutes:icpp:2018,puri:40gbcharmodels:arxiv:2018}. Prior work also explores other forms of data parallelism~\cite{krizhevsky:weirdtrickcnnparallel:arxiv:2014, recht:hogwild:nips:2011,maleki2018semantics}, and there are opportunities to reduce data communication overheads of parallelism~\cite{alistarh:gradquantization:nips:2017,wen:gradquantization:nips:2017,lin:gradcompression:iclr:2018}. Layer parallelism can also reduce the memory requirements for model weights~\cite{shen:cnnacceleratorpartitioning:isca:2017, shen:escher:fccm:2017}. Improved model parallelism techniques could recover some of the \textasciitilde23\% algorithmic FLOP utilization lost to layer parallelism in the case study. Frameworks should aim to automatically and dynamically subdivide the computation, automatically map appropriate compute graph portions to compute resources, and prefetch data between host and accelerator memories.

\textbf{Operational Intensity:} Our modeling also shows that RNN networks suffer from moderate operational intensity in large matrix multiplications. This medium operational intensity is caused by the need to stream inputs from memory multiple times during tiled multiplies, decreasing the algorithmic FLOP utilization by \textasciitilde40\% compared to an ideal system that would not need to restream inputs. In this setting, increasing on-chip cache size may hinder compute throughput growth, but is likely to proportionally reduce input restreaming from memory. Better cache tiling, kernel optimization and fusion techniques might also help~\cite{coates:cotshpc:icml:2013,chetlur:cudnn:arxiv:2014}.

Hardware techniques to better support large-scale RNN training---larger memories and on-chip caches---run counter to emerging accelerator designs. Emerging designs aim to support very high compute-to-memory ratios by optimizing for compute throughput. This design philosophy is unlikely to trade die area for memory channels (capacity) or caches.

%% file: related.tex
\section{Other Related Work}\label{sec:related}

\noindent\textbf{DL Application Characterization:}
Many benchmark suites have been developed that aim to analyze DL applications by focusing on particular ops/kernels~\cite{deepbench} or quantifying the performance of end-to-end DL applications~\cite{dawnbench,mlperf:benchmark:2018}. OpenAI recently characterized the trend in DL FLOP growth over time using coarse approximations of FLOPs for a range of applications ranging from AlexNet to Alpha Go~\cite{openai}.

\noindent\textbf{Performance Modeling}
Paleo is an analytical performance model which explores parallelism for CNN networks~\cite{qi:paleo:iclr:2017}. 




%
 

%% file: conclusion.tex
\section{Conclusion}
This paper leverages the prior work to project the dataset and model size growth required to advance DL accuracy beyond human-level, to ``frontier'' targets defined by machine learning experts. Datasets will need to grow by $33$--$971\times$, while models will need to grow by $6.6$--$456\times$ to achieve target accuracies. We project the computational requirements to train these applications at scale. Our results reveal an important segmentation of DL training challenges for recurrent neural networks (RNNs) that contrasts with prior studies of deep convolutional networks. RNNs will have comparatively moderate operational intensities and very large memory footprint requirements. In contrast to emerging accelerators, large-scale RNN training characteristics suggest designs with significantly larger memory capacity and on-chip caches.

%% file: appendices.tex
\onecolumn
\section{Artifact Appendix}
\label{sec:appendix_artifact}

\subsection{Abstract}
\textit{The artifact contains the latest version of our codebase, called ``Catamount'', which is a compute graph analysis tool to load, construct, and modify deep learning (DL) models and to symbolically analyze their compute requirements. Catamount can read DL model checkpoints saved from DL frameworks (e.g., from Tensorflow). This artifact includes (A) Tensorflow checkpoints for each of the models (compute graphs) analyzed in the PPoPP 2019 paper, \textbf{Beyond Human Level Accuracy: Computational Challenges in Deep Learning} and (B) shell scripts to run graph analytics and generate results for Figures 7 to 10 of the paper. To validate the results, run the test script, which generates and collects the corresponding outputs:}

\begin{verbatim}
    ~$ bash catamount/frameworks/example_graphs/tensorflow/full_models/generate_results.sh
\end{verbatim}

\subsection{Artifact check-list (meta-information)}

\begin{itemize}
  \item \textbf{Algorithm:} Catamount can construct or load compute graphs and using various graph traversal algorithms, propagate symbolic graph dimensions and calculate various compute requirements, including compute Flops, memory accesses, and memory footprint.
  \item \textbf{Data set:} Consists of compute graph checkpoints from neural network models trained in Tensorflow. The models are a word language model (LSTM), character language model (RHN), neural machine translation (encoder/decoder+attention), speech recognition (encoder/decoder+attention), and image classification (ResNet).
  \item \textbf{Operating system:} Linux
  \item \textbf{Hardware:} No special hardware is required. CPU-based system. Recommended 8+ GB memory.
  \item \textbf{Program requirements:} Python 3.6
  \item \textbf{Python packages:} Catamount requires Python packages to run. First, it requires a virtual environment created with Virtualenv (\url{https://virtualenv.pypa.io/en/latest/}, commonly included with Python 3.6), or another virtual environment package. Second, Catamount depends on three other Python packages, \textit{numpy}, \textit{sympy}, and \textit{tensorflow>=1.7}. See instructions below to install these dependencies.
  \item \textbf{Input:} Each Catamount test takes as input the Tensorflow model definition (compute graph) for the problem domain. Model descriptions are in Section~\ref{sec:applications}.
  \item \textbf{Output:} Catamount tests output all analytics about the models (compute graphs) they load, including symbolic model parameters, algorithmic Flops, memory accesses, and memory footprint. By binding these symbolic functions to particular values, the tests also output the numerical values.
  \item \textbf{How much disk space required (approximately)?:} 1 GB
  \item \textbf{How much time is needed to set up experiments (approximately)?:} Less than 5 minutes
  \item \textbf{How much time is needed to complete experiments (approximately)?:} Less than 2 hours, depending on system CPU, memory performance
  \item \textbf{Publicly available?:} Yes
  \item \textbf{Artifact DOI:} \url{https://doi.org/10.5281/zenodo.2259280}
  \item \textbf{Repository location:} \url{https://github.com/baidu-research/catamount}
  \item \textbf{Code/data licenses (if publicly available)?:} Apache 2.0
\end{itemize}

\subsection{Description}

\subsubsection{How delivered}
Catamount is an open source Python package under Apache 2.0 license and is hosted with code and example DL compute graphs on GitHub (\url{https://github.com/baidu-research/catamount}).

\subsubsection{Software dependencies}
Catamount depends on recent versions of \textit{numpy}, \textit{sympy}, and \textit{tensorflow>=1.7}, which work most stably with recent versions of Python. It is strongly recommended that users begin with Python 3.6 to install and run Catamount tests. Catamount may not work smootly with prior versions of Python.

\subsubsection{Data sets}
The model definitions (compute graphs) analyzed in this paper are included as end-to-end tests of Catamount functionality and are distributed in the Catamount Github repository. These model definitions are in the form of Tensorflow checkpoints that were saved using the standard Tensorflow saver as follows:

\begin{verbatim}
    # ... Construct a TF model ...
    # Set output directory
    outdir = ...
    # Start TF session, create saver, and save model
    with tf.Session() as sess:
        sess.run(tf.global_variables_initializer())
        saver = tf.train.Saver()
        saver.save(sess, os.path.join(outdir, 'tf_graph'))
\end{verbatim}

\noindent
This process saves the graph definition as a Tensorflow MetaGraphDef file, \texttt{tf\_graph.meta}, along with saved parameters. Catamount can load the MetaGraphDef (\texttt{.meta}) files as graph-like Python objects for analysis. An example saving process can be found in the Catamount repo at \texttt{catamount/frameworks/example\_graphs/tensorflow/rnn/tf\_dyanmic\_rnn.py}. 

The \texttt{.meta} files used to analyze compute requirements for the different applications in this paper can be found in the following locations in the repo:
\begin{itemize}
    \item Machine translation, word and character language models: \newline \texttt{catamount/frameworks/example\_graphs/tensorflow/full\_models/language\_models/}
    \item Image classification ResNet models: \newline \texttt{catamount/frameworks/example\_graphs/tensorflow/full\_models/image\_classification/}
    \item Speech recognition attention model: \newline \texttt{catamount/frameworks/example\_graphs/tensorflow/full\_models/speech\_attention/}
\end{itemize}

Finally, the tests that load these graphs and generate analytical models and numerical outputs can be found in the Catamount full-graph tests directory:
\begin{itemize}
    \item Language models: \texttt{catamount/tests/full/tf\_language\_models.py}. Pass the parameter \texttt{--domain <domain>}, where \texttt{<domain>} can be one of \texttt{charlm}, \texttt{nmt}, or \texttt{wordlm}, for character, machine translation, or word models, respectively.
    \item Image classification: \texttt{catamount/tests/full/tf\_image\_resnet.py}. Pass the model depth as a parameter \texttt{--depth}. Supported depths currently include ResNet 18, 34, 50, 101, or 152.
    \item Speech recognition: \texttt{catamount/tests/full/tf\_speech\_attention.py}
\end{itemize}

\subsection{Regenerating experiments from this paper}
To download Catamount and regenerate results for this paper, run the following commands. These commands clone the public repository, check out the commit known to work for validating PPoPP paper results, and run the tests that generate results:

\begin{verbatim}
    ~$ git clone https://github.com/baidu-research/catamount
    ~$ cd catamount
    ~$ git checkout -b ppopp-artifact-validation ppopp-2019-artifact
    ~$ bash catamount/frameworks/example_graphs/tensorflow/full_models/generate_results.sh
\end{verbatim}

\subsection{Evaluation and expected result}
With the commands in the prior subsection, Catamount should create an output file for each of the 9 analyzed compute graphs. These files will be named \texttt{ppopp\_2019\_tests/output\_*.txt}. To regather results from these files after running the \texttt{generate\_results.sh} script, you can run the following command inside the top-level Catamount directory:
\begin{verbatim}
    ~$ bash catamount/frameworks/example_graphs/tensorflow/full_models/gather_results.sh
\end{verbatim}

\subsection{Experiment customization}
To customize Catamount tests or experimental results from this paper, users can modify the appropriate Python script in the tests directory, \texttt{catamount/tests/full/}. By changing the the \texttt{bind\_subs} dictionary, users can bind symbolic dimensions of the compute graphs to different values. Users can also create their own compute graph definitions manually using the Catamount API (see \texttt{catamount/api/}) or by checkpointing Tensorflow models and loading them into Catamount as shown in the tests.

Catamount can calculate the algorithmic compute requirements for a loaded model. In particular, it calculates algorithmic FLOPs, memory bytes accessed, and minimal memory footprint for a pass through the compute graph. In addition to requirements presented in the paper, Catamount can calculate \textbf{algorithmic IO}, which is the amount of data accessed for input to and output from a model. Training data is often stored on and read from disk, and placed into the model's input memory allocations. Algorithmic IO is proportional to the batch size, but stays fixed as model size and training step compute requirements grow. We do not investigate algorithmic IO in this work, because we expect IO will grow very slowly relative to compute.